\documentclass[letterpaper, 10pt, conference]{ieeeconf}      % Use this line for a4 paper
\IEEEoverridecommandlockouts                              
\usepackage{graphicx}
\usepackage{tikz}
\usepackage{lipsum}
\usepackage{mathtools}
\usepackage{cuted}

\usetikzlibrary{positioning,arrows}
\pgfdeclarelayer{background}
\pgfdeclarelayer{foreground}
\pgfsetlayers{background,main,foreground}
\usepackage{footmisc}
\usepackage{caption}
\usepackage{subfigure}
\usepackage{amsmath}
\usepackage[version=4]{mhchem}
\usepackage{siunitx}
\usepackage{longtable,tabularx}
\usepackage{tabularx}
\usepackage{multicol, blindtext}
\usepackage{multirow}
\usepackage{xcolor}
\usepackage{hyperref}
\usepackage{tikz}
\usetikzlibrary{shapes,arrows}

\tikzset{block/.style={draw, thick, text width=2cm ,minimum height=1.3cm, align=center}, line/.style={-latex}     }  

\IEEEoverridecommandlockouts                              % This command is only needed if 
\title{\LARGE \bf Flow Synthesis Based Visual Servoing Frameworks for Monocular Obstacle Avoidance Amidst High-Rises}
\author{Harshit K. Sankhla*$^{1}$, M. Nomaan Qureshi*$^{1}$, Shankara Narayanan V.*$^{1}$, Vedansh Mittal$^{1}$, Gunjan Gupta$^{1}$,\\  Harit Pandya$^{2}$, K. Madhava Krishna$^{1}$
\thanks{*indicates equal contribution, orders determined by dice rolling.}% <-this % stops a space
\thanks{$^{1}$Robotics Research Center, IIIT Hyderabad, India}%
\thanks{$^{2}$Cambridge Research Laboratory, Toshiba Europe, UK}%
}

\begin{document}
\maketitle

\begin{abstract}
We propose a novel flow synthesis based visual servoing framework enabling long-range obstacle avoidance for Micro Air Vehicles (MAV) flying amongst tall skyscrapers. Recent deep learning based frameworks use optical flow to do high-precision visual servoing. In this paper, we explore the question: can we design a surrogate flow for these high-precision visual-servoing methods, which leads to obstacle avoidance? We revisit the concept of saliency for identifying high-rise structures in/close to the line of attack amongst other competing skyscrapers and buildings as a collision obstacle. A synthesised flow is used to displace the salient object segmentation mask. This flow is so computed that the visual servoing controller maneuvers the MAV safely around the obstacle. In this approach, we use a multi-step Cross-Entropy Method (CEM) based servo control to achieve flow convergence, resulting in obstacle avoidance. We use this novel pipeline to successfully and persistently maneuver high-rises and reach the goal in simulated and photo-realistic real-world scenes. We conduct extensive experimentation and compare our approach with optical flow and short-range depth-based obstacle avoidance methods to demonstrate the proposed framework's merit. Additional Visualisation can be found at \url{https://sites.google.com/view/monocular-obstacle/home}
\end{abstract}

\section{Introduction}

As UAVs gain popularity, interest has grown in the development of robust navigation methods. For the urban setting, navigation through high-rises has become a problem of imminent interest. Low-cost off-the-shelf RGBD sensors are noisy \cite{CCO-Voxel} and detect obstacles in the typical stereo depth range, not efficient for long-range obstacle detection and maneuvering. In this context, we propose a novel flow synthesis based deep visual servoing framework for monocular obstacle avoidance, wherein by obstacles, we indicate urban high-rises/skyscrapers. Independently and otherwise, flow-based approaches to monocular avoidance \cite{Sensors, SENSORS1} have been popular in literature; nonetheless, their outcomes have tended to be unreliable due to often inaccurate flow estimates and controllers based on such flow. Instead, by synthesizing a desired flow, we preempt the challenges associated with the noisy flow, and the multi-step CEM-based control achieves superior goal convergence than a single-step reactive control based on flow. 

\begin{figure}
\begin{center}
\includegraphics[width=0.9\linewidth]{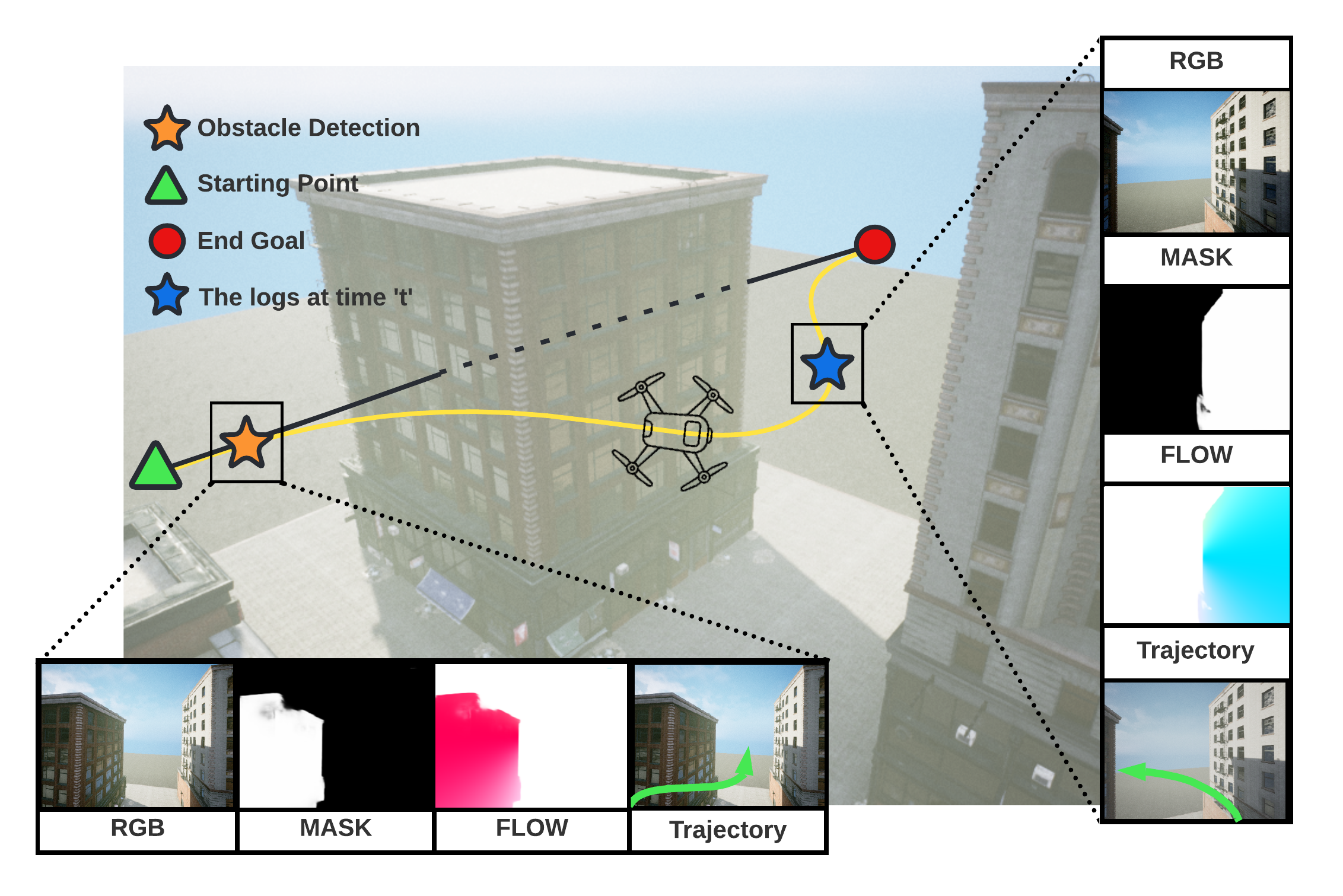}
\end{center}
\caption{Our algorithm explicitly segments the building to generate a surrogate desired flow for the high precision Flow-Based visual servoing algorithms. The 'pink' obstacle flow pushes the building leftward, and hence a rightward drone trajectory is generated by the servoing framework and vice-versa for 'blue' flow, which pushes the building rightward.}
\label{fig:teaser}
\end{figure}

Recent Visual Servoing frameworks \cite{ICRA-20,CoRL-20, IROS-21} have leveraged deep optical flow to do high precision visual servoing. In this paper, we ask the question: If these frameworks can utilise optical flow to reach a goal image with high precision, can we design a surrogate flow for the obstacle that can push the obstacle out of the scene? Given an urban high-rise scene as a monocular image input, our novel pipeline segments the Building of Concern (BoC) through a modified saliency segmentation network. This network segments the BoC amongst other competing high-rises. The BoC is typically the building that will collide first with the MAV if it continues along its current trajectory. The pixels contained in the segmentation mask of the BoC are subject to a flow that quintessentially is the pixel error that needs to be minimized by the multi-step visual servoing based CEM controller. Fig. \ref{fig:teaser} showcases a high-level overview of our idea. 
% The central idea is to mimic the flow pattern or flow trajectory, an obstacle is subject to, when a camera performs an avoidance maneuver around that obstacle. The optical flow sequence of an obstacle as the camera maneuvers around it is shown in figure \cite{FlowPatternA}. The net displacement of the pixels of the obstacle due to the avoidance maneuver (see figure \cite{FlowPatternA}) is typically the synthesized flow imparted to the obstacle. Hence when the multi-step CEM controller strives to achieve this synthesized flow or minimizes this flow error, it realizes obstacle avoidance (figure \cite{FlowPatternA}). 

To the best of the authors' knowledge, this approach is novel yet uncomplicated. It parts ways from prior flow-based methods that synthesized a control corresponding to the observed flow of an obstacle. Rather, in this case, a flow error is synthesized that is devoid of any noise and easily reduced by a controller. Moreover, unlike data-driven frameworks that directly learn and map pixels to control often through black-box neural networks, the proposed CEM controller offers sufficient interpretability of what it strives to accomplish.

\subsection{Contributions}

The paper contributes in the following ways:
\begin{enumerate}
    \item To the best of the authors' knowledge, this is the first approach to tackling the problem of navigation amongst urban high-rises with a single monocular camera as the essential sensing modality.
    \item We revisit the concept of salient object detection as collision obstacle detection. This deep model segments only the Building of Concern (BoC), thereby ensuring the relevance of our servoing based avoidance mechanism for an identified obstacle.
    \item In contrast to previous methods of synthesizing control from estimated flow, which is often noisy, we propose noiseless surrogate flow synthesis and demonstrate its efficacy for monocular obstacle avoidance.
    % \item The paper benchmarks tangible performance gain with respect to popular flow-based obstacle avoidance methods \cite{FlowAvoid} and other prior art when evaluated on photo-releastic simulation environments in AirSim.
    % \item In the context of UAM, the paper showcases the importance of image-based flow synthesis as an enabler of long-range avoidance maneuvering and portrays its advantage vis a vis short-range depth-based obstacle avoidance pipelines \cite{FastPlanner}.
\end{enumerate}
The paper benchmarks tangible performance gain in comparison with popular flow-based obstacle avoidance methods \cite{SENSORS1}, and short-range depth-based obstacle avoidance pipelines \cite{FastPlanner} on both photorealistic and real-world scenes imported into the AirSim Simulation Environment \cite{AirSim}.
% The efficacy of the proposed framework is vindicated by successful goal attainment in a variety of urban scenes, both photorealistic and real-world scenes imported into the AirSim Simulation Environment \cite{AirSim} based on Unreal Engine. 

\section{Related Work}

\textbf{Monocular Obstacle Avoidance:} While the literature is sufficiently populated with methods relating to obstacle avoidance with depth sensors such as RGBD cameras \cite{FastPlanner} or stereoscopy \cite{Voxblox, EdgeFlow}, prior art relating to monocular obstacle avoidance has been sparse. In \cite{Cremers} a Monocular Direct SLAM framework popularly called LSD-SLAM is used to construct occupancy maps over which navigation and exploration is performed. However, monocular SLAM based occupancy maps are typically noisy that entail uncertainty-based obstacle avoidance formulation along the lines of \cite{CCO-Voxel} as an overhead. Moreover, as the authors mentioned, maps often need to be regenerated by intermittent hovering maneuvers that can prove costly as UAVs and MAVs work with limited onboard power and compute budget. In \cite{Hamid} a similar SLAM-based approach using feature-based ORB SLAM is presented. The sparse map is interpreted in terms of planes, and an RRT-based trajectory planner is presented. Once again, such sparse point cloud methods are subject to uncertainty-based overheads and need extra modules beyond SLAM to reinterpret the sparse point clouds.

\textbf{Optical Flow and Feature Based Avoidance Methods:} In contrast to methods that explicitly build a map, there exists a class of methods that make use of optical flow cues to compute obstacle avoidance behavior. Flow balancing is a popular method of obstacle avoidance where the difference in flow between the horizontal and vertical sections of the image are balanced by inducing an appropriate yaw and height changes in the vehicle \cite{Sensors, SENSORS1}. Elsewhere flow is used to compute depth or time to collision that guides the control performance \cite{ComputationalBiology}. In \cite{ComputationalBiology} correlation-based motion detectors were also tried out as a mechanism for obstacle detection taking cues from biology. Whereas in \cite{IV-16} growth in feature sizes were the cue to detect the presence of an obstacle, and heuristic controls were used for avoidance. A primary challenge with optical flow-based control lies in the flow estimates being noisy. Often successive flow estimates can cause conflicting control actions such as back and forth motion primarily due to inaccuracies or noisy flow estimates. 

\begin{figure*}[ht!]
\begin{center}
\includegraphics[width=13cm, height=7.0cm]{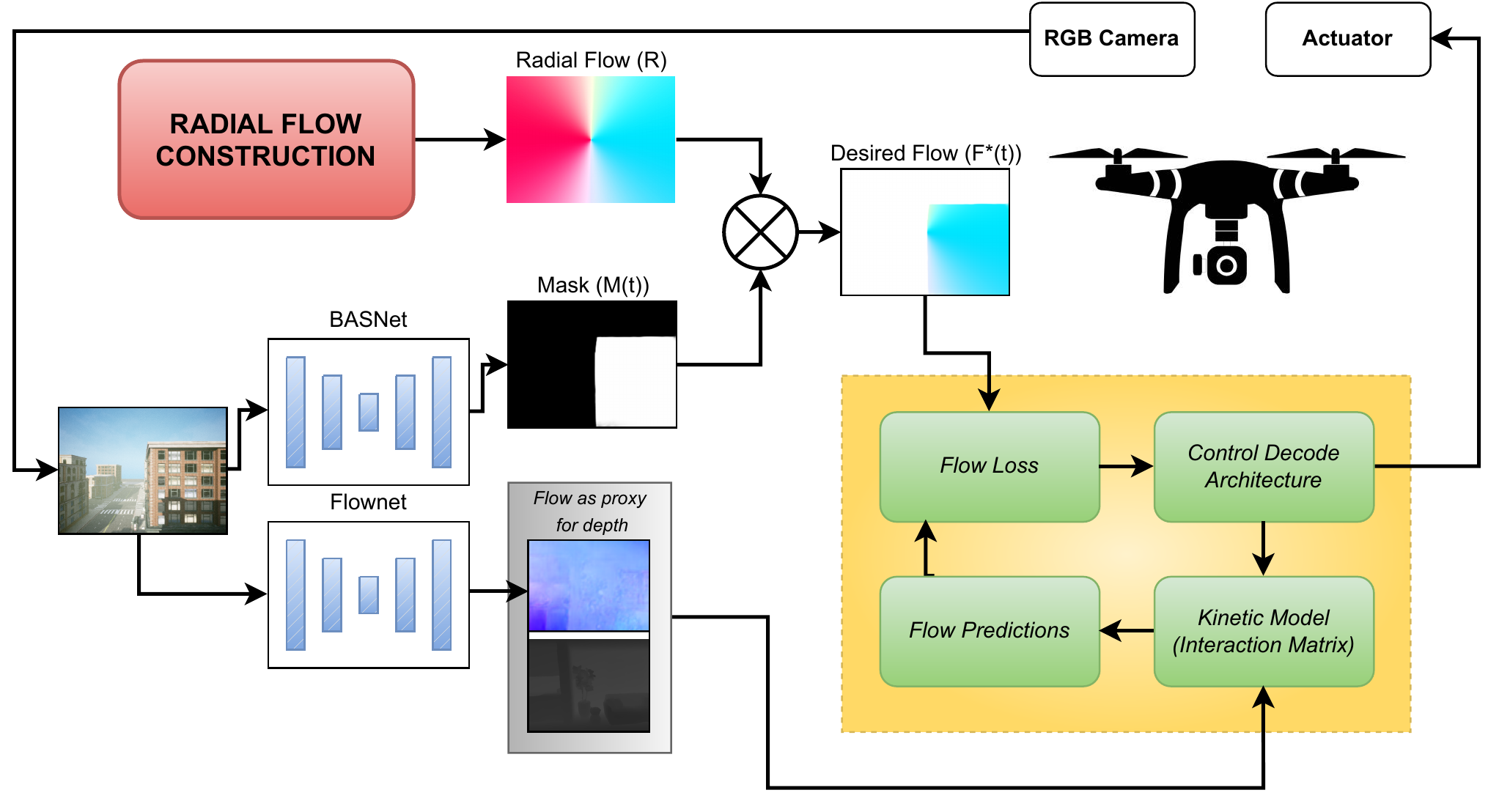}
\caption{
Here we summarize our obstacle avoidance pipeline. We use the saliency-based detection module BASNet to get the obstacle mask $\mathcal{M}(t)$. We then combine it with the radial flow $\mathcal{R}$ to get the desired flow $\mathcal{F}^*(t)$. We use the synthesized desired flow $\mathcal{F}^*(t)$ along with a flow-based Visual Servoing pipeline to generate a control command using Cross Entropy Method. 
} 
\label{fig:pipeline}
\end{center}
\end{figure*}

\textbf{Data Driven Methods:} Data-driven methods for monocular obstacle avoidance have also been popular. Some paradigms use black-box neural nets to learn disparity \cite{Tinne} followed by a traditional controller to avoid the perceived obstacles over the depth maps; other methods directly map and infer pixels to controls \cite{Dronet, sergey}. Depth maps learned from monocular vision are noisy and need further retraining in new scenes and entail that the controller resorts to uncertainty-based formulations for obstacle avoidance such as in \cite{CCO-Voxel}.

\textbf{Visual Servoing:} This is an active camera control technique that derives a control command to achieve the pose for a target view, with respect to the current viewpoint \cite{Hutchinson}. Literature abounds with a variety of servoing frameworks \cite{vs_manip1,vs_manip2} while recently data-driven methods have gained in popularity \cite{ICRA-20, CoRL-20, IROS-21}. This paper uses a multi-step CEM-based servoing controller that outputs control and minimizes synthesized flow errors.

\textbf{Contrast with Previous Methods:} Unlike the mapping formulations, the present approach operates directly on images than on the estimates of the 3D world from the images, thereby circumventing problems associated with noisy maps and the sparseness of the reconstruction. Synthesizing the control corresponding to flow estimated over images has been the prevalent practice, such as in optical flow-based and depth-based avoidance methods \cite{Sensors, SENSORS1, Tinne}. However, the present method synthesizes the flow and generates control that achieves this synthesized flow. Since the synthesized flow is noiseless as obtained by construction, the CEM Servoing Controller that minimizes this synthesized flow error is untroubled largely by uncertainty in the flow estimates. Further, unlike the reactive approaches to flow-based obstacle avoidance \cite{Sensors} the DeepMPC controller introduced in \cite{CoRL-20} is a controller over a time horizon that provides increased robustness. We show vastly superior performance over reactive optical flow-based avoidance controller under the duress of flow noise. Further, this paper is believably the first attempt to explore the role of visual servoing in an obstacle avoidance setting, whereas all previous approaches have used servoing to achieve a goal-reaching behavior (reaching the pose corresponding to the desired image)

\section{Method}
Given a camera and a GPS sensor, our aim is to generate the optimal velocity control commands to reach a desired goal GPS location $g^*$ in a collision-free manner. Modern Deep Learning-based servoing methods \cite{ICRA-20, CoRL-20} use optical flow for high precision visual servoing. We ask the question: if these methods can use optical flow to reach a precise goal image, can we design a surrogate flow for obstacles that can push them out of the image?
Referring to Fig.\ref{fig:pipeline}, we first utilise the saliency-based architecture described in \cite{Qin_2019_CVPR} to identify the obstacle or building in the line of attack of the MAV. We then introduce the Radial Flow in Sec. \ref{secradial}, which can be combined with the obstacle mask obtained using BASNet to provide a desired flow for the servoing frameworks introduced in \cite{ICRA-20, CoRL-20}. We use MPC+CEM \cite{CoRL-20} to optimise for velocity commands; however, this only serves as the design choice.

\subsection{Saliency Based Obstacle Avoidance}
\label{saliency}
Saliency detection methods \cite{borji2015salient} aim to mimic the human visual attention mechanism to identify objects more attentive than the surroundings. Saliency detection has a range of applications, such as foreground-background segmentation, object tracking, identification, reidentification, detection, especially in cluttered environments. A relevant yet sparsely explored application is collision obstacle detection during navigation. A MAV traversing in an urban environment must navigate many structurally diverse high-rises at different times during the flight. To identify any such 'Building of Concern', we leverage the power of Salient Object Detection models and find them to perform well in detecting obstacles even at long distances. Further, our downstream avoidance task utilises the generated saliency map to synthesise flow and generate velocity commands that can steer the drone clear from a collision. It can also efficiently work with imperfect object segmentation masks; hence, salient object detection methods justify our case for detecting objects of collision.

The salient object detection method we are using is BASNet\cite{Qin_2019_CVPR}, a U-Net\cite{ronneberger2015u} like supervised encoder-decoder based method with an additional residual refinement module which improves the coarse prediction, extending it to the object boundary. We train BASNet to identify obstacles that appear in the line of sight of collision. Typical semantic segmentation methods \cite{9356353} would segment all competing object instances in the entire field of view without any criteria such as the object's location. Depth-based methods like \cite{FastPlanner} are not reliable beyond stereo camera ranges and would detect large obstacles like buildings very late as compared to when they are visible. Notably, Salient object detection methods handle the task well for long-range collision obstacle detection from a single RGB image.

 We train BASNet on a curated dataset of MAV FPV images of skyscrapers from around the world and model buildings in UrbanScene3D and AirSim (building 99 environment). We perform data augmentation to increase the dataset size to 4000, with transformations such as \textit{ColorJitter, MotionBlur, RandomBrightness, RandomRotate} and follow the training methodology of the original authors, using the same hyperparameters. 
% where the initial learning rate=1e-3, betas=(0.9, 0.999), eps=1e-8, and weight decay=0. The training loss converges after 400 iterations on a computer with an 8-core AMD Ryzen 7 2700X CPU, 64 GB RAM, and an Nvidia GTX 1070 Ti GPU.

% Imagine a drone trying to navigate from one point to another. Although multiple obstacles might lie in its path, at any given point of time, only one obstacle might be of particular interest. We train the BASNet\cite{basnet} to this cause to generate a mask $M(t)$ at time-step "t". 

\subsection{Radial Flow}
\label{secradial}
Recent visual servoing methods \cite{ICRA-20, CoRL-20} calculate the flow between the current image and desired image to do high precision visual servoing. However, during the monocular obstacle avoidance, the desired image is unavailable. Here we explain how we can design a surrogate flow, which can help us robustly avoid obstacles. Consider an image of dimension $(H \times W)$. For any pixel coordinate $(i, j)$, we want to assign a flow value that pushes the pixel out of the image. This turns out to be a flow in a radially outward direction from the image center. This ``radial" flow $R$ and for any pixel coordinate can be described using :
\begin{align}
\label{eq:radial }
\textstyle R[i, j] &= [i - \frac{H}{2}, \Lambda*(j - \frac{W}{2})]
\end{align}

Where $\Lambda$ is a parameter and is set to 10. $\Lambda$ remains same for each scene in our benchmark. We normalize the Radial Flow as: 
\begin{align}
\label{eq:NR}
\textstyle N[i, j] &= \sqrt{(i - \frac{H}{2})^2 + (j - \frac{W}{2})^2)} \\
\textstyle R[i, j] &= \frac{R[i, j]}{N[i,j]}
\end{align}

We use the mask $M(t)$ obtained using our saliency based obstacle avoidance module to get the obstacle mask. This mask $M(t)$ is then multiplied with Normalised Radial Flow $R$ to get a desired flow for servoing as described in the next section. 
\begin{align}
\label{eq:desiredflow}
\textstyle \mathcal{F}^*(t) &= R \cdot M(t) 
\end{align}

\begin{figure}[ht!]
\centering
\includegraphics[width=0.8\linewidth]{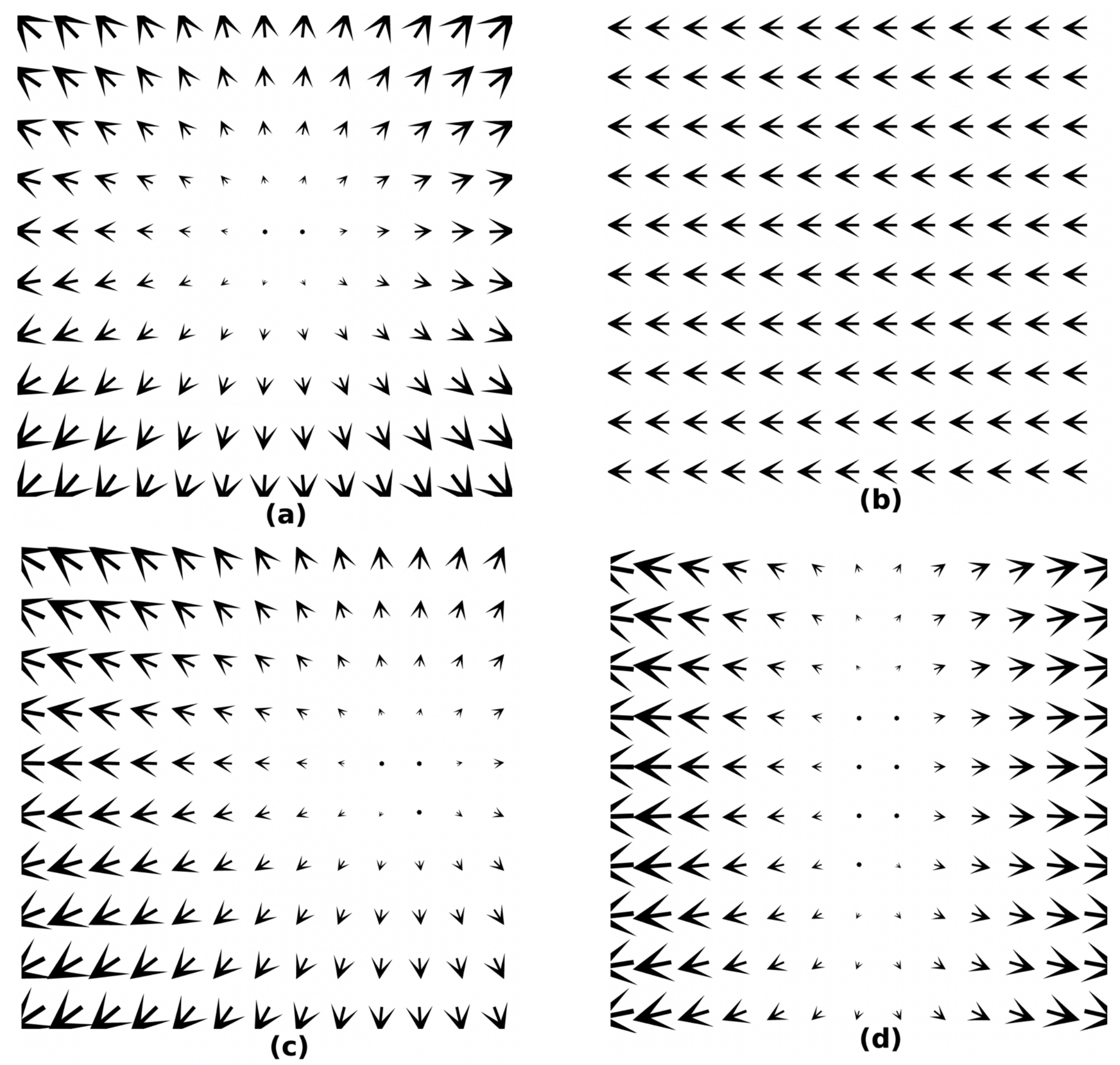}
  \caption{\textbf{Visualisation of flow vectors according to how the drone moves :} (a), (b) show the flow obtained when the drone moves forward and right, respectively. (c) shows the flow when the drone moves forward and right simultaneously. (d) shows a visualisation of the synthesized radial flow using Eq.\ref{eq:radial }. We leverage the fact that when (d) is combined with the obstacle mask $\mathcal{M}(t)$, it forces the CEM optimisation to generate a velocity that moves the obstacle out of the scene. So if the mask shows that the building is dominantly in the right direction, the servoing generates the velocity, which takes it in (forward+left) direction, towards the side of the building, and vice-versa if the mask is dominantly on the left side.}   
\label{fig:radialflow}
\end{figure}

The motivation behind the design of the radial flow can be explained using Fig.\ref{fig:radialflow}. We want the desired flow as calculated in Eq. \ref{eq:desiredflow} to be biased to take the MAV away from the obstacle while also taking the MAV forward. However, we cannot assign the pixel-wise optical flow in a radially outward direction since it is the same as when the drone moves forward. Hence, we scale the component along the (left, right) direction, forcing the optimisation to produce a velocity that takes it away from the building while also moving forward. Additionally, if the center patch of the image is covered with an obstacle, we damp the forward velocity by a factor $\mu$. This provides a robust surrogate desired flow $\mathcal{F}^*(t)=R*\mathcal{M}(t)$ which can be used with a flow-based servoing algorithm to reliably generate optimal velocity commands as explained in section \ref{secavoidance}. 

% \begin{figure}[h!]
% \begin{center}
% \includegraphics[width=1.0\linewidth]{Figures/radialFlow.png}
% \end{center}
%   \caption{Qualitatively understanding the radial flow : For every pixel coordinate $(i, j)$, we assign it a flow which pushes it in a radially outward direction. Instead of generating a whole new scene and generating a flow for that, we utilise the approximate optical flow obtained to induce a control law. }
% \label{radialflow}
% \end{figure}

\subsection{Avoidance Using Visual Servoing Framework}
\label{secavoidance}
We now explain our algorithmic pipeline, which combines the saliency mask predicted using BASNet and the radial flow to give a velocity control command to avoid the obstacles. The salient obstacle mask is obtained using the BASNet module. The mask at time $t$ is combined with Radial Flow $R$ to give a desired flow $F^*(t)$ for the current scene using Eq. \ref{eq:desiredflow}. We then calculate the 2-View Flowdepth $Z_T$ introduced in \cite{ICRA-20} which considers the optical flow between time step $t-1$ and $t$ as a proxy estimate for depth. We use the flow network FlowNet2\cite{flownet2} to construct the interaction matrix $L(Z_t)$.
\begin{equation}
\label{eq:interaction}
L(Z_t) = \begin{bmatrix*}[r]
    \frac{-1}{Z_t} & 0 & \frac{x}{Z_t} & xy  & -(1+{x}^2) & y \\
    0 & \frac{-1}{Z_t} & \frac{y}{Z_t} & 1+{y}^2  & -xy & -x
    \end{bmatrix*}.
\end{equation}

The interaction matrix $L(Z_t)$ maps the drone velocity to the velocity of pixels (or optical flow) on the image plane. We generate the flow predictions using the interaction matrix $L(Z_t)$. 
\begin{equation}
\label{eq:predictive_model}
%  \begin{aligned}
 \begin{split}
 \widehat{\mathcal{F}}(\mathbf{v}_{t+1:t+T}) %1/T\sum_{k=1}^{T} [\mathcal{{F}}_{pseudo}(I_{t+k},I_{t+k+1})]\\
 = \sum_{k=1}^{T}[L(Z_{t})\mathbf{v}_{t+k}]
 \end{split}
%  \end{aligned}
\end{equation}

We can then optimize the velocity for the desired flow:
\begin{equation}
\label{eq:flow_loss}    
\mathcal{L}_{flow} = \lVert \widehat{\mathcal{F}}(\widehat{\mathbf{v}_{t}}) - \mathcal{F}^{*}\rVert = \lVert [L(Z_{t})\widehat{\mathbf{v}}]
 - \mathcal{F}^*\rVert
\end{equation}

Optimising the loss function in Eq. \ref{eq:flow_loss} gives the desired velocity for the avoidance of obstacle. We use Cross-Entropy Method (CEM) to solve for the desired velocity. At each time step we sample 'N' velocities from a Gaussian distribution ${{g}(\mu, \sigma^{2})}$ and calculate the loss (Eq. \ref{eq:flow_loss}) for each of them. The losses are then sorted, and the top 'K' velocities (with least losses) are used to update the parameters of Gaussian distribution ${{g}(\mu, \sigma^{2})}$. We run CEM for several iterations before convergence is achieved. This is analogous to the MPC+CEM described in \cite{CoRL-20}.

\subsection{Overall Pipeline and Implementation Details}
We now describe our overall goal-reaching pipeline. We divide our algorithm into two parts, a) Goal Reaching mode: Here, we use GPS location to orient and move towards the goal location $g^*$. b) Obstacle avoidance mode: We give the velocity commands obtained (as explained in Section \ref{secavoidance}) to the drone. We use the obstacle mask $\mathcal{M}(t)$ obtained using our Saliency Based Obstacle detection algorithm to decide whether to follow Obstacle avoidance mode or not. If the center patch of the image is covered with an obstacle, we damp the forward velocity by a factor $\mu$. Our method can be used to do 6-DOF obstacle avoidance, however, we used a 4-DoF ($x, y, z, yaw$) while presenting the results in this paper. 

\subsection{Flow Balancing with Radial flow}
In this section, we describe how we can use the desired radial flow $\mathcal{F}$ obtained using Eq. \ref{eq:desiredflow} to improve the performance of the Flow Balancing Method \cite{Sensors}. Flow balancing has been classically used to avoid obstacles using a monocular camera. It uses the flow between the current image and the previous image to get a yaw angle for the drone. The yaw rate is given using the equation : 
\begin{equation}
\label{eq:flow_balance}    
\mathcal{\dot{\theta}} = 
(\frac{\omega_L - \omega_R}{\omega_L + \omega_R}) \times k
\end{equation} 

Here $\omega_L$ and $\omega_R$ represent the sum of magnitudes in the left and right half of the Optical flow between the current image $I_t$ and previous image $I_{t-1}$ and k is a constant which is an adjustable parameter. The reasoning behind the approach is that closer obstacles have higher flows, so we should move to the direction that has less flow. However, this method fails to work on our benchmark because, in long-range avoidance, the optical flow fails to provide enough information to give an optimal velocity command. To have a fair comparison against flow-balancing, we use the formulated radial flow $F^*(t)$ described in (Eq. \ref{eq:desiredflow}) to get the yaw rate. Our Radial Flow-based Flow balancing gives much better results when compared to naive flow balancing. 

\begin{figure*}[h!]
\begin{center}
\includegraphics[width=0.8\linewidth]{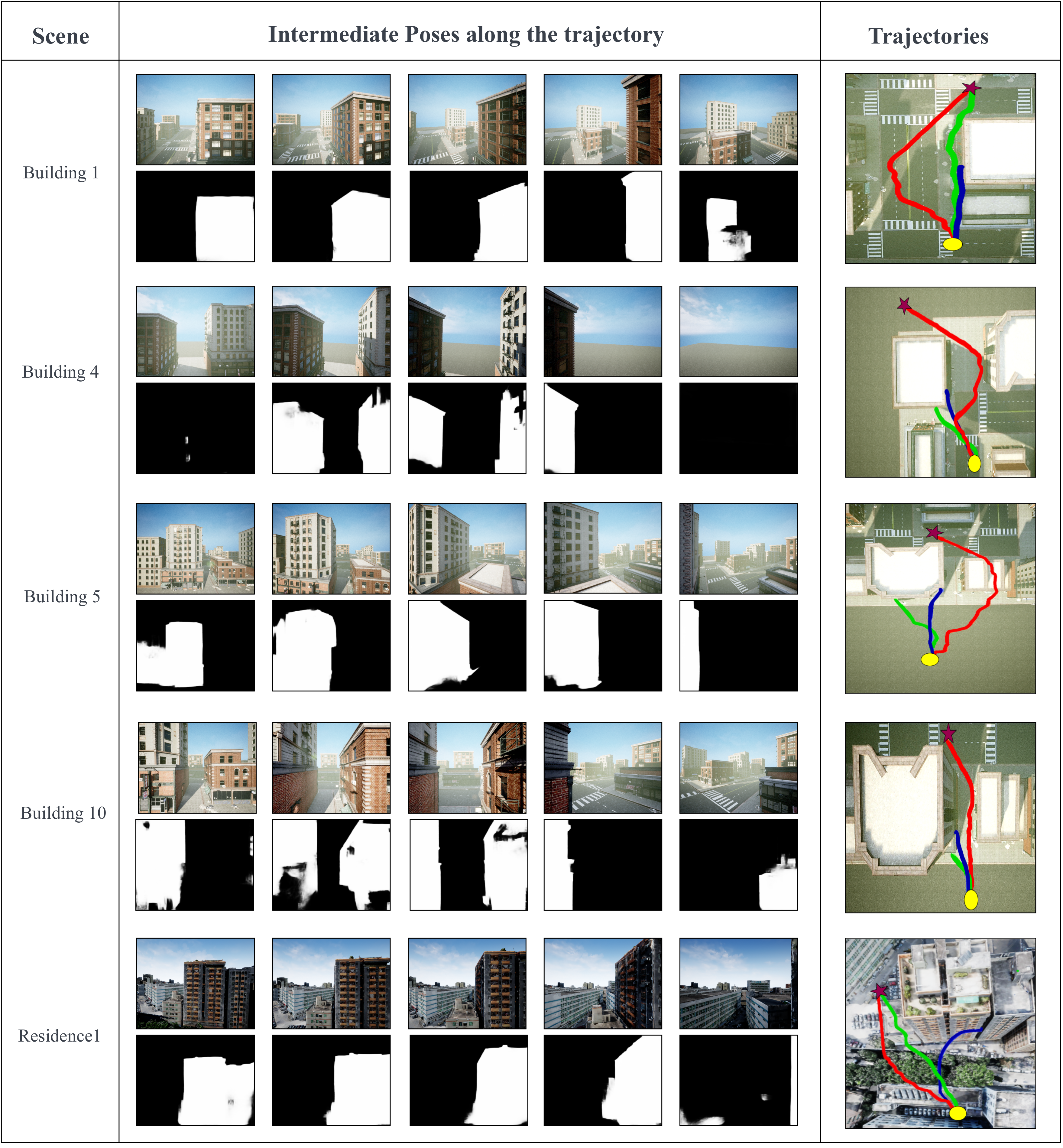}
\end{center}
  \caption{\textbf{Qualitative results on the benchmark for selected scenes}: Our method successfully avoids the obstacles on all the 10 scenes in the building 99 environment and 5 of 6 in the UrbanScene3D environments. Here we show images for 5 intermittent poses captured during the obstacle avoidance for selected scenes in the simulation benchmark. In Building 1, we are able to segment and avoid the building even though there are several buildings behind it. Building 4 covers a large part of the image, but our algorithm is able to move in the correct direction. In Building 10, the MAV can navigate the narrow path between the two buildings successfully. We also present results on certain challenging configurations from the real-world dataset UrbanScene3D.
The lines in the trajectory column indicate the following: (RED) --$>$ Our Method, (BLUE) --$>$ Flow Balancing, (GREEN) --$>$ Flow Balancing with Radial Flow. The goal and start positions are marked with a red star and a yellow circle, respectively. 
  }
\label{fig:quali_table}
\end{figure*}

\section{Experiments}
\begin{table*}[htbp]
\centering
% {p{0.2\textwidth}>{\centering}p{0.2\textwidth}>{\centering}p{0.2\textwidth}>{\centering\arraybackslash}p{0.2\textwidth}}
\fbox{\begin{tabular}{|p{0.1\textwidth}||>{\centering}p{0.1\textwidth}|>{\centering}p{0.1\textwidth}||>{\centering}p{0.1\textwidth}|>{\centering}p{0.1\textwidth}||>{\centering}p{0.1\textwidth}|>{\centering\arraybackslash}p{0.1\textwidth}|}
\hline
 Scenes & \multicolumn{2}{c||}{Naive Flow Balance}& \multicolumn{2}{c||}{Radial Flow based Flow Balance}&  
 \multicolumn{2}{c|}{Our Method} \\
\cline{2-3} \cline{4-5} \cline{6-7} 
        &  Min Dist  & Traj Length &  Min Dist  & Traj Length
	&  Min Dist  & Traj Length    \\    \hline
%Shankara -  a '-' instead of collision	
$Building1 $ & $\times$ & $\times$ & 4.9801 & 65.733  &  \textbf{8.2812} & 78.651  \\ \hline
% $Building2 $ &  $\times$ & $\times$ & $\times$ & $\times$  & \textbf{8.4453} & 89.152  \\   \hline     
% $Building3 $ &  $\times$ & $\times$ & 0.2807 & 81.730  & \textbf{1.3222} & 58.425  \\  \hline
 $Building4 $ & $\times$ & $\times$ & $\times$ & $\times$  & \textbf{3.4189} & 98.542 \\\hline
 $Building5 $ & $\times$ & $\times$ & $\times$ & $\times$  & \textbf{9.6503} & 96.384  \\   \hline   
%  $Building6 $ & $\times$ & $\times$ & $\times$ & $\times$  &  \textbf{8.7850} & 74.818  \\\hline
%  $Building7 $ & $\times$ & $\times$ & $\times$ & $\times$  & \textbf{12.273} & 79.469  \\\hline
% %  BUILDING 15 ON CODE  = BUILDING 9 HERE
%  $Building8 $ & $\times$ & $\times$ & $\times$ & $\times$  & \textbf{2.9671} & 77.462  \\    \hline
%   $Building9 $ & 1.2713 & 50.617  & 3.6186 & 50.186  & \textbf{7.4885} & 53.414  \\ \hline
 $Building10 $ & $\times$ & $\times$  & $\times$ & $\times$ & \textbf{3.0070} & 59.624 \\    \hline
%  $Building1 $ & 0.1004 & $\times$ & 4.9801 & 65.733  &  \textbf{8.2812} & 78.651  \\ \hline
% $Building2 $ &  0.1265 & $\times$ & 0.4623 & $\times$  & \textbf{8.4453} & 89.152  \\   \hline     
% $Building3 $ &  0.2131 & $\times$ & 0.2807 & 81.730  & \textbf{1.3222} & 58.425  \\  \hline
%  $Building4 $ & 0.5264 & $\times$ & 0.1009 & $\times$  & \textbf{3.4189} & 98.542 \\\hline
%  $Building5 $ & 0.1002 & $\times$ & 0.2214 & $\times$  & \textbf{9.6503} & 96.384  \\   \hline   
%  $Building6 $ & 0.5871 & $\times$ & 0.3774 & $\times$  &  \textbf{8.7850} & 74.818  \\\hline
%  $Building7 $ & 0.4028 & $\times$ & 0.3451 & $\times$  & \textbf{12.273} & 79.469  \\\hline
% %  BUILDING 15 ON CODE  = BUILDING 9 HERE
%  $Building8 $ & 0.3757 & $\times$ & 0.3421 & $\times$  & \textbf{2.9671} & 77.462  \\    \hline
%   $Building9 $ & 1.2713 & 50.617  & 3.6186 & 50.186  & \textbf{7.4885} & 53.414  \\ \hline
%  $Building10 $ & 0.3571 & $\times$  & 0.1437 & $\times$ & \textbf{3.0070} & 59.624 \\    \hline

 $Residence1 $ & $\times$ & $\times$ & 0.1072 & 38.636  & \textbf{11.818} & 66.058  \\   \hline 
%  $Residence2 $ &  $\times$ & $\times$ & $\times$ & $\times$  & \textbf{11.937} & 123.748 \\   \hline 
%  $Sci-Art1 $ & 6.6581 & 108.258 & 13.007 & 108.386 & \textbf{25.484} & 103.597  \\  \hline           
%  $Sci-Art2 $ & $\times$ & $\times$ & $\times$ & $\times$  & \textbf{15.039} & 167.726  \\   \hline 
%  $Campus1 $ & $\times$ & $\times$ & $\times$ & $\times$  & \textbf{10.265} & 108.532  \\  \hline          
%  $Campus2 $ & $\times$ & $\times$ & $\times$ & $\times$  &  $\times$ & $\times$  \\  \hline  
 
%  $Residence1 $ & 0.6571 & $\times$ & 0.1072 & 38.636  & \textbf{11.818} & 66.058  \\   \hline 
%  $Residence2 $ &  0.1561 & $\times$ & 0.1900 & $\times$  & \textbf{11.937} & 123.748 \\   \hline 
%  $Sci-Art1 $ & 6.6581 & 108.258 & 13.007 & 108.386 & \textbf{25.484} & 103.597  \\  \hline           
%  $Sci-Art2 $ & 0.1409 & $\times$ & 0.2836 & $\times$  & \textbf{15.039} & 167.726  \\   \hline 
%  $Campus1 $ & 0.1604 & $\times$ & 0.1765 & $\times$  & \textbf{10.265} & 108.532  \\  \hline          
%  $Campus2 $ & 0.2552 & $\times$ & 0.1043 & $\times$  &  0.1029 & $\times$  \\  \hline 
 
 \hline
 
 \end{tabular}}
\caption {\label{tab:mindistance} \textbf{Minimum Distance and Trajectory Length: } Min Dist signifies the minimum distance of MAV from any building in the image. We show that our method has the highest Min Dist among all the methods, which empirically proves that our method maintains the safest distance from the buildings. It is clear from Fig. \ref{fig:quali_table} trajectory plots that our method takes the safest path to the goal while other methods either collide with the building or graze past it, and hence our trajectory length is highest. "$\times$" indicates that the maneuver was not completed, and the drone collided.}
\end{table*}
% \endgroup

\subsection{Simulation Benchmark}
Our simulation benchmark consists of 10 photo-realistic scenes from the building 99 environment and 6 real-world reconstruction scenes from the UrbanScene3D dataset \cite{UrbanScene3D}. These scenes span across a varying difficulty level depending on the number of buildings on the path to the goal location, the amount of free space available to the MAV while navigating between buildings, and the angle at which it approaches the building. 
% We have a start and a random goal GPS position for each scene. 
% In easy scenes, the robot has to follow a straight-forward trajectory to reach the desired goal location, while the medium scenes and hard scenes involve reaching the goal location after navigating across buildings by following a complicated trajectory. Additionally, hard scenes are designed to have space constraints, such that the agent has to make some clever decisions while navigating in narrow spaces.  
%The difference between a hard scene and a medium scene involves the density of buildings. It hence relates to the constraints on space available for the drone to navigate while avoiding buildings.  
The urban scene 3D dataset contains real-world urban environments for cities like Shanghai, New York, etc., with realistic textures. This benchmark further tests the generalization of our methods by putting them in near-realistic scenes. We have chosen to conduct our experiments on the real scenes: Sci-Art, Residence, and Campus. Our method is able to navigate to the provided goal GPS location in all the scenes and is able to make intelligent choices required to navigate in scenes with space constraints. Fig. \ref{fig:quali_table} shows the obstacle avoidance sequence for several scenes on our benchmark. 

We show that our algorithm is able to navigate to the goal location in scenes across varying levels of difficulty. For Building 1, BASNet segments only the building in the line of attack amongst multiple competing buildings. For Building 4 and Building 10, the algorithm has to navigate in a narrow passage between two skyscrapers and make several key decisions to reach the respective desired goal locations. We are also able to navigate in scenes where a large part of the scene is covered with an obstacle. In Building 5, our visual servoing optimisation pipeline is able to find the correct direction to move in order to avoid the obstacle. On the photo-realistic UrbanScene3D dataset, our algorithm generalises and gives similar performance. Our algorithm is able to work on 15 out of 16 scenes. Compared to our proposed novel algorithm, naive Flow Balancing works on only 2 scenes out of 16 scenes from our simulation benchmark. It is clear that the optical flow between the current and previous image is not able to provide any meaningful direction for the MAV to follow. However, our radial-flow guided flow-balancing algorithm shows a stark improvement compared to naive flow balancing. Still, it fails to generalise and can successfully reach goal location only on 5 out of 16 scenes in our simulation benchmark. 

The success-rate results are summarised in  Table \ref{tab:srate}, and it is clear that our method shows robust performance and is able to generalise across the Building-99 and UrbanScene3D datasets. Table \ref{tab:mindistance} shows the minimum distance of the camera from any building. Our algorithm consistently performs better than other algorithms and maintains the highest safe distance from the obstacles. This is also evident from Fig. \ref{fig:quali_table} where we plot the trace of trajectories obtained using different algorithms. 

\begin{table}[h!]
\begin{center}

\setlength\tabcolsep{0.5pt}
\begin{tabular}{|l|l|l|l|}
\hline
\multicolumn{1}{|l|}{Approaches} & \begin{tabular}[c]{@{}c@{}}UrbanScene3D\end{tabular} & \begin{tabular}[c]{@{}c@{}}Building 99\end{tabular} & \begin{tabular}[c]{@{}c@{}}Total\\Success Rate \end{tabular} 
\\ \hline
\hline
Naive  Flow-Balancing \cite{SENSORS1} & 1/6 & 1/10 & 12.5\%
\\ \hline
Radial-Flow + Flow-Balancing  & 2/6 & 3/10 & 31.25\%
\\ \hline
\textbf{Ours} & \textbf{5/6}* & \textbf{10/10}* & \textbf{93.75\%}*
\\ \hline
\end{tabular} 

\caption{\label{tab:srate} \textbf{Success Rate Comparison}: Our method generalises across different scenes in the Building-99 and UrbanScene3D dataset to give a consistent controller performance.}
\label{fig:success_rate}
\end{center}
\vspace{-0.2cm}
\end{table}

% \subsection{Quantitative Results}
% We have compared our method with Naive Flow Balancing and Radial Flow based Flow balancing. We report both the qualitative and quantitative results of our method on the benchmark. It can be noticed that Naive Flow Balancing was able to avoid on x scenes in easy and failed in all the medium and hard scenes. The Radial Flow based Flow balancing showed 
% \subsection{4DOF Avoidance}

\subsection{Salient Object Detection as Collision Obstacle Detection}
\begin{figure}[ht!]
\centering
\includegraphics[width=0.9\linewidth]{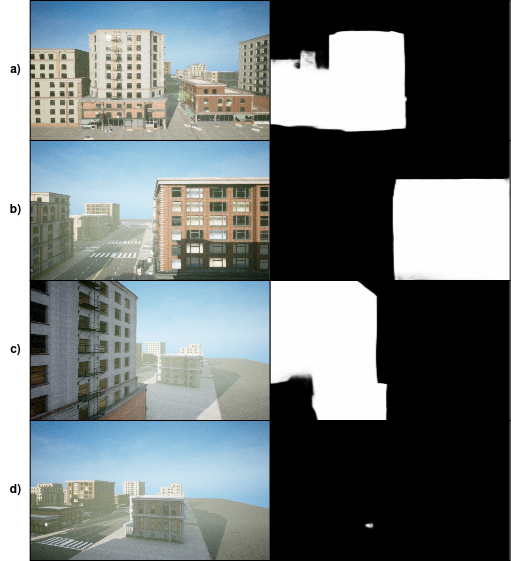}
  \caption{Obstacle segmentation masks for scenes in the AirSim Building 99 environment computed from BASNet}
%  \vspace{-0.7em}
\label{fig:saliency}
\end{figure}

In this section, we present qualitative results of BASNet in detecting the BoC during flight. Fig. \ref{fig:saliency}a)-d) show detection of obstacles in/close to the line of attack of the drone, except in Fig. \ref{fig:saliency}d) owing to its small size and distance from the drone. However, it must be noted that this is not a collision scenario, and such behavior is actually desirable so the drone does not remain in a perpetual state of detecting and avoiding an obstacle and can continue moving on the original trajectory. Fig. \ref{fig:saliency}a) and c) also show how imperfect/partial segmentation masks do not hinder downstream flow synthesis and subsequent avoidance, as the velocity computed from the desired flow still moves the drone away from the obstacle.

\begin{figure}[ht!]
\centering
\includegraphics[width=0.9\linewidth]{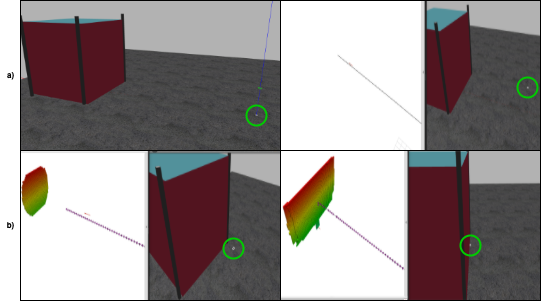}
  \caption{Evaluation of stereo depth avoidance for a large structure. The MAV is enclosed in a white bounding box, position highlighted with a green circle. \textbf{a)} shows the simulation setup in Gazebo consisting of a building with a 30mx30m front and an Iris drone approaching it from 30m distance and 10m altitude. \textbf{b)} shows that the detection of a building (voxel map) starts when the MAV is 9m away from the front and ends up being 1m close till the whole front is detected. The planner fails to find a trajectory to avoid the obstacle; subsequently, the MAV is halted.}
 \vspace{-1em}
\label{fig:stereoavoid}
\end{figure}

\subsection{Stereo Depth based Avoidance}
We compare the performance of our obstacle detection and avoidance method with a stereo depth based avoidance method. We use Fast-Planner\cite{FastPlanner} a kinodynamic path searching method using a 3D point cloud as perception information on a MAV in a simulated environment in Gazebo and approaching a building 30m wide and 60m tall. This method works on voxelised occupancy mapping for environment representation, making it agnostic to the scene/object texture. Fig. \ref{fig:stereoavoid} shows the experiment setup, obstacle occupancy map, and MAV trajectory. We highlight the result in terms of obstacle detection range and subsequent failure for avoidance. Stereo range based depth cameras work reliably in the range of 7-10m, which is essentially how close the MAV gets to the building before seeing it as an obstacle. At this point, for a typically wide building, it becomes unfeasible to view around the sides and plan any path for avoidance, resulting in the drone halting in its track with no further course of action other than the nondeductive exploration of a vast structure to find a safe passage. This becomes inefficient for MAVs by wasting limited flight endurance.

\section{Conclusion}
This paper presented a robust monocular obstacle avoidance algorithm deployed for UAVs navigating amongst urban environments. Our framework leverages saliency-based segmentation to identify the Building of Concern. We then combine the obtained segmentation mask with the proposed radial flow to get the desired flow, which can be fed into the high precision flow-based visual servoing methods to avoid the obstacles successfully. We conduct extensive experiments to demonstrate the merit of our approach. Our work shows how a visual servoing framework and a saliency-based obstacle detection can be combined to avoid obstacles robustly. In future work, we aim to explore saliency and visual servoing based frameworks to avoid obstacles at a higher velocity. 

\bibliographystyle{IEEEtran}
\bibliography{IEEE}

\end{document}